# Patient level simulation and reinforcement learning to discover novel strategies for treating ovarian cancer

Brian Murphy, M.S.[1]
Mustafa Nasir-Moin, A.B.[2]
Grace von Oiste[2]
Viola Chen, M.D.[3]
Howard A Riina, M.D.[2,4,5]
Douglas Kondziolka, M.D.[2,6]
Eric K Oermann, M.D.[2,4,7]*

[1] Vilcek Institute of Graduate Biomedical Sciences, NYU Grossman School of Medicine, New York, NY 10016
[2] Department of Neurosurgery, NYU Langone Health System, New York, NY 10016
[3] Oncology Early development, Merck & Co., Inc, Kenilworth, NJ 07033, USA.
[4] Department of Radiology, NYU Langone Health System, New York, NY 10016
[5] Department of Neurology, NYU Langone Health System, New York NY 10016
[6] Department of Radiation Oncology, NYU Langone Health System, New York NY 10016
[7] Center for Data Science, New York University, New York, NY 10016
*Corresponding Author

Corresponding Author:
Eric Oermann, MD
Assistant Professor of Neurosurgery, Radiology, and Data Science
NYU Langone Health
530 First Ave
Skirball, 8R
New York, NY 10016
eric.oermann@nyulangone.org

## ABSTRACT

**Purpose**: The prognosis for patients with epithelial ovarian cancer remains dismal despite improvements in survival for other cancers. Treatment involves multiple lines of chemotherapy and becomes increasingly heterogeneous after first-line therapy. Reinforcement learning with real-world outcomes data has the potential to identify novel treatment strategies to improve overall survival. We design a reinforcement learning environment to model epithelial ovarian cancer treatment trajectories and deploy a reinforcement learning agent to suggest therapeutic regimens for simulated patients.

**Experimental Design**: We construct a Markov decision process that serves as a reinforcement learning environment using Cox proportional hazard survival analysis and patient-level data from The Cancer Genome Atlas. We use a temporal difference learning approach to solve for optimal treatment strategies based on simulated patient trajectories from the environment. We test average survival time of patients treated by the agent at the start and end of training and against real patients treated by clinicians. We also measure the frequency and timing of the agent's treatments and compare them with clinician decisions.

**Results**: The agent identifies treatment strategies that improve average survival within the conditions of the environment. However, it prefers experimental drugs and is not able to find a stable policy when restricted to more common actions.

**Conclusions**: This experiment demonstrates proof-of-concept for a reinforcement learning approach to identifying epithelial ovarian cancer treatment strategies and can serve as a prototype for future efforts. The work would benefit from a larger sample size of real-world cancer patients and refinement of the Cox survival models.

**Statement of Translational Relevance**: Evolutionary dynamics and evolved treatment resistance is a hallmark of disseminated cancer and is often responsible for eventual cancer progression and patient death. Typically multiple lines of single or multi-agent therapies are administered over time in an attempt to counter these dynamics of resistance while minimizing toxicity and maximizing overall survival. The advances in knowledge of this study include: (1) the construction of a patient-level simulation of response to individual lines of therapy for metastatic ovarian cancer using real-world outcomes data; (2) the creation of a game whereby an agent selects lines of therapy for simulated patients in order to maximize a reward signal (overall survival); (3) a model-free reinforcement learning agent that learns to find an optimal solution for this game.

# INTRODUCTION

While research has led to marked increases in survival rates for many cancers, the dismal prognosis for epithelial ovarian cancer patients (approximately 47% five-year survival) has not improved substantially over the years.(1–3) Treatment regimens for epithelial ovarian cancer typically involve surgical debulking followed by adjuvant chemotherapy, which often proceeds over several lines of therapy with each line using multiple agents in combination.(1,2,4–8) Typical first-line treatment for epithelial ovarian cancer involves combination of a platinum based compound such as carboplatin and a taxane based compound such as paclitaxel.(2,6,9) Physicians may further optimize these lines of treatment by combining chemotherapy with immunotherapy or radiation, adjusting dosing schedules, and using different drug administration routes.(6,10) As patients progress into additional lines of therapy, overall treatment plans become increasingly heterogeneous with little guidance for optimizing towards overall survival.(1,2,6,9) These complex and heterogeneous regimens used in practice present a challenge for attempts to develop personalized optimal care pathways based on published literature.(11,12) An alternative approach could be to learn optimal treatment plans directly from trial and registry data conditioned on individual patient characteristics.

Reinforcement learning is a branch of machine learning that uses artificial intelligence (AI) algorithms, "agents," to discover optimal strategies for achieving a goal. Reinforcement learning agents function within an environment which can consist of discrete states that describe the environment to the agent, and administer numerical rewards which the agent then tries to maximize via the actions it takes. (13,14) Given its ability to learn and optimize algorithms for sequential decision making, reinforcement learning offers potential benefits in clinical decision making for diseases that involve multiple treatment stages. (15–22) In this study, we reformulate real world cancer trial data as a patient-level simulated environment of epithelial ovarian cancer and train a reinforcement learning agent to predict optimal treatment decisions for patients.

# METHODS

*Data Source and Preprocessing*

609 patients with epithelial ovarian cancer were retrieved from The Cancer Genome Atlas ("TCGA"). Comprehensive treatment plans and responses to therapy for these patients were obtained from Villalobos et al., 2018.(7) We employed a pre-processing pipeline (see link to code) to prepare the dataset prior to conversion into a reinforcement learning environment. We converted all drug names to their generic equivalent names using a drug standardization index based on the NCI Drug Dictionary and Broad GDAC Firehose (see, https://gdisc.bme.gatech.edu/cgi-bin/gdisc/tap5.cgi).(23) We removed samples from the treatment regimens data that did not include the names of the drugs in the treatment line, and where the treatment start and end days were equivalent, making the timing of the treatment line unclear. We additionally excluded patients that did not achieve their overall survival endpoint leaving us with 225 out of 460 patients in the final dataset. (Table 1).

Next, we reorganized the data into 30-day treatment periods. The full reorganized dataset consists of 9,296 one-month treatment period samples, each containing the patient ID, the number of months since the start of treatment, and the current combination of therapeutic agents, including periods where the patient was not prescribed any treatment. These data contain 127 unique drug combinations and a "no active treatment" option. We used the subset of patients whose final survival metric was a death event to construct a reinforcement learning environment. This subset consists of 5,931 one-month treatment period samples, with 107 unique drug combinations and the "no active treatment" option.

*Environment: State*

These data were then used to construct a Markov decision process (MDP) to simulate the treatment decisions and survival trajectories of epithelial ovarian cancer patients. This served as the reinforcement learning environment in the model, with each state constituting the overall status of the patient, their response to the current line of therapy, as well as time since the start of treatment, total time on therapy, age, race, and tumor specific information (tumor grade and stage). The actions available to the agents consisted of all unique treatment combinations (see Supplemental Methods for detailed environmental description). We additionally excluded drug combinations not present in the TCGA ovarian cancer dataset due to insufficient information.

*Environment: Survival modelling*

The critical element of our environment is modeling a set of transition probabilities of each patient from one state to the next given a particular treatment at a particular time. Each state transition involves two sets of probabilities that stochastically determine the subsequent state. The first determines whether a patient dies with probability $P(D)$ or survives with probability $P(S) = 1 - P(D)$. If the patient dies, the subsequent state is death, the terminal state, and simulation proceeds to the next round. If the patient survives, the next set of probabilities determines whether she goes into remission, $P(R)$, or needs further treatment in the next state, $P(T) = 1 - P(R)$ (Supplemental Figure S1). To calculate these probabilities, we used a survival analysis based on two multivariate Cox proportional hazard regressions. The first regression calculates baseline hazard using a terminal death event and months since the start of treatment, and the second calculates baseline hazard using the recurrent remission event and months on the current treatment regimen (Supplemental Figure S2). Each regression calculates treatment-specific hazard with the feature set that describes the patient state and the actions from each state. We then sample each regression's survival function according to the patient's current state-action pair to obtain $P(D)$, $P(S)$, $P(R)$ and $P(T)$ (see Supplemental Methods for detailed description of survival modeling). The reward is the total number of months of patient survival where the action $a$ did not result in death.

*Reinforcement learning and statistics*

We opted for a model free reinforcement learning approach and implemented a deep Q-network (DQN) (see Supplemental Methods for details). The agent chooses actions (drug combinations) based on observed state transitions, and the state-action pairs feed into the MDP which stochastically determines subsequent states (Figure 1A, Supplemental Figure S1).

The DQN agent was trained for 200,000 rounds where one round = one simulated patient, and previous patient trajectories served as the training dataset for the DQN (Figure 1B). During training, we calculated the cumulative average survival and the moving average survival of the previous 1,000 simulated patients. Final performance for the DQN agent was evaluated based on the average simulated patient survival of the last 1,000 patients of training compared to a baseline average survival calculated from the first 1,000 patients. We also compared the average survival of the last 1,000 simulated patients with the average survival of patients treated by clinicians (i.e., patients with overall survival in the TCGA dataset). We used a two-sided t-test to test the significance of the average survival at the end of training compared to each baseline average at an alpha of 0.05. We characterized treatment decisions for clinicians and the AI policy by calculating the relative frequency that each drug combination occured at fixed time intervals from the start of treatment. For heatmaps displaying these patient treatment characteristics, chemotherapy administration frequencies were normalized from time of initial treatment and converted to z-scores across all drug combinations. Z-scores were subsequently bounded between 0-3 to generate smooth visualizations of drug selection and timing in courses of care by both the DQN and human oncologists. We also calculated the percentage of patients receiving each drug combination in each line of treatment, displaying values for the first six treatment lines from the TCGA data, the DQN agent, and the restricted DQN.

We separately ran the experiment restricting the actions available to the DQN agent to the drug combinations that occured at least five times in the dataset used to construct the MDP. This "restricted" model was intended as a control on the agent to prevent it from learning actions drawn from rare drug combinations given in unique circumstances that may not be reflective of general oncological care. Lastly, we designed a separate, rules-based agent based on the National Comprehensive Cancer Network (NCCN) guidelines for ovarian cancer treatment and deployed it in the same environment to simulate performance based on available medical guidance.(9)

**RESULTS**

After running the reinforcement learning simulation with the full set of actions for 200,000 simulated cases, we noted a marked shift in policy after approximately 50,000 simulated patients (Figure 2A), whereby the agent learned to use aldesleukin resulting in an increase in mean survival from 32.3 months for the first 1,000 patients to 42.9 months for the last 1,000 patients (p<0.000) (Figure 2A-B). The AI treatment strategy also exceeded the baseline oncologist driven treatment strategy, which had a mean survival of 26.4 months, for both the first 1,000 (p=0.003) and last 1,000 patients (p<0.000, Figure 2B). Notably, human oncologists most commonly prescribed carboplatin and paclitaxel as first-line treatment and over time switched to topotecan, doxorubicin, carboplatin, or paclitaxel monotherapy in later months (Figure 4A, Figure 5A), while the DQN favored an almost continuous administration of aldesleukin (Figure 4B, Figure 5B).

We also ran the simulation with the restricted action set to see whether the DQN agent would develop an alternative strategy when limited to only more commonly used treatments. After a million simulated cases, the DQN developed a policy that favored gemcitabine and tamoxifen combination therapy and shifted to other regimens such as cisplatin and tamoxifen combination therapy in later rounds (Figure 4C, Figure 5C). After training, the restricted DQN was able to significantly improve average survival, with a mean survival of 45.5 months for the last 1,000 simulated patients compared to the restricted baseline (first 1,000 patients) of 43.4 months (Figure 3A-B, p=0.099). This also represented a statistically significant increase over the average clinician, with p-values less than 0.000 for the two-sided t-test between the clinicians and both the baseline and trained agent (Figure 3B). However, though the restricted DQN seemed to slowly improve its average performance through approximately 450,000 rounds of training, it is not clear that it settled on a stable policy after a million rounds, as did the unrestricted DQN (Figure 3A).

For an additional comparison, we ran the simulation with the NCCN agent for 200,000 rounds and calculated the average survival across all simulated patients. The NCCN agent outperformed clinicians (p<0.000) and the last 1,000 patients for both the DQN (p<0.000) and restricted DQN (p<0.000) with average survival of 49.5 months.

**DISCUSSION**

We present the first use of a reinforcement learning agent to create novel, individualized treatment plans for cancer patients after training on a patient level simulator of cancer outcomes based on real-world data. Previous studies have explored the use of reinforcement learning to select specific dosing regimens for single agents (20,21,24), to optimize radiotherapy plans (25), and to optimize artificial therapies in simulated clinical trials (26). However, prior works fail to account for the wide variety of available treatments, and their potential for impacting outcomes when administered in a sequential manner as part of a comprehensive oncological treatment plan. In order to accomplish this, we pioneer the use of real-world data to develop a patient level simulation of epithelial ovarian cancer treatment and outcomes to serve as a simulation environment for training a DQN agent to the task of optimizing treatment plans. Previous approaches towards treatment plan discovery and optimization have focused on searching the existing clinical trial literature, an approach popularized by IBM Watson (11,12), and from genomic and molecular biomarker data (27–30). These static approaches, applied at a single time point, fail to account for the evolution of cancer in response to treatments (1) and the goal of maximizing overall survival across an entire course of care.

Interestingly, our NCCN rules based agent obtained the overall best results as measured by overall survival outcomes. This is an encouraging validation of our simulation and the NCCN guidelines demonstrating that overall adhering to the guidelines lead to the best overall survival for our simulated cohort. Validating guidelines using data-driven simulations of cancer is itself an interesting implication of this work. Guidelines have evolved over the past several decades with increasingly sophisticated treatments and a corresponding improvement in overall survival (31,32). An interesting future direction and further validation of this cancer simulation technology could be to see if the simulated patients experience similar survival trends when treated under the same set of evolving guidelines over time.

A further sensible future direction would be to test our models on patients that were not used to create the training environment that a pre-trained agent would "treat" by observing their characteristics as it did with simulated patients and recommending treatments. The reinforcement learning agent would need to prove successful in this task to be considered for clinical use. In a clinical setting, the agent could inform oncologists' decisions for real patients using data from the patient's medical records and test results as inputs for treatment suggestions at each phase of their cancer and treatment trajectory.

This paper has several key limitations worth noting. First and foremost, the quality of our trained agent is fundamentally limited by the fidelity of the underlying simulator. We constructed our patient-level cancer simulation on the TCGA dataset annotated by Villalobos et al. (7), and only included patients who achieved their overall survival endpoint limiting our overall sample size. This high dimensional dataset with a relatively small sample size makes for a particularly challenging reinforcement learning problem, and a clear future direction is to improve upon the existing simulation with a larger amount of real world data curated from one or more comprehensive cancer centers or clinical trial databases. A sufficiently massive dataset would be at least an order of magnitude larger in terms of the number of patients (225 in this study), and ideally two orders of magnitude larger, on par with the MIMIC-III dataset that Komorowski et al. used for their model.(17) A further, critical limitation of our model is the mathematical model of survival. Results from the simulation, specifically the fact that the untrained DQN agents outperformed clinicians in average survival (Figure 2B, Figure 3B), show that the survival model requires further improvements to more accurately represent the cancer treatment environment. We employed multivariable Cox proportional hazard regression (see Supplemental Methods for details), but further refinements of our survival model and a comprehensive model of toxicity would likely lead to marked improvements in overall performance.

In conclusion, we present the first use of real world data to create a patient level simulation of cancer for the training of an AI agent to optimize individual care pathways for patients suffering from epithelial ovarian cancer. Although primarily a proof of concept, we expect that the combination of cutting edge AI technologies such as reinforcement learning with real-world data has the potential to significantly contribute to basic oncological decision making at the point of care in clinical scenarios that extend beyond current guidelines.

# TABLES
**Table 1:**

| Table 1 \| Data Summary | | |
|---|---|---|
| Demographics, tumor characteristics, survival outcomes | | |
| | All Patients | Deceased Only |
| Unique patients (n) | 460 | 225 |
| Age, years (mean (s.d)) | 58.7 (11.3) | 59.9 (10.8) |
| Final survival outcome (n (%)) | | |
|     Living | 235 (51.1) | n/a |
|     Deceased | 225 (48.9) | 225 (100.0) |
| Overall survival, days (mean (s.d.)) | 1017.6 (765.8) | 1141.1 (690) |
| Race (n (%)) | | |
|   White | 403 (87.6) | 205 (91.1) |
|   Black or African American | 20 (4.3) | 12 (5.3) |
|   Asian | 15 (3.3) | 3 (1.3) |
|   American Indian or Alaska Native | 2 (0.4) | 2 (0.9) |
|   Native Hawaiian or other Pacific Islander | 1 (0.2) | 1 (0.4) |
|   Not specified | 19 (4.1) | 2 (0.9) |
| Tumor Stage (n (%)) | | |
|   IIIC | 332 (72.2) | 166 (73.8) |
|   IV | 67 (14.6) | 37 (16.4) |
|   IIIB | 20 (4.3) | 10 (4.4) |
|   IIC | 19 (4.1) | 5 (2.2) |
|   IIIA | 7 (1.5) | 4 (1.8) |
|   IC | 6 (1.3) | n/a |
|   IIB | 3 (0.7) | 2 (0.9) |
|   IIA | 2 (0.4) | n/a |
|   IB | 2 (0.4) | n/a |
|   IA | 1 (0.2) | n/a |
|   Not specified | 1 (0.2) | 1 (0.4) |
| Tumor grade (n (%)) | | |
|   G3 | 391 (85) | 187 (83.1) |
|   G2 | 54 (11.7) | 30 (13.3) |
|   GX | 8 (1.7) | 4 (1.8) |
|   G1 | 5 (1.1) | 2 (0.9) |
|   GB | 1 (0.2) | 1 (0.4) |
|   Not specified | 1 (0.2) | 1 (0.4) |

**Table 1:** Overall survival is the number of days to the later of death, last follow up, last recorded tumor progression or recurrence, and last recorded chemotherapy. Only patients with a final survival endpoint ('Deceased Only') were included in the analyses in this study.

**A**

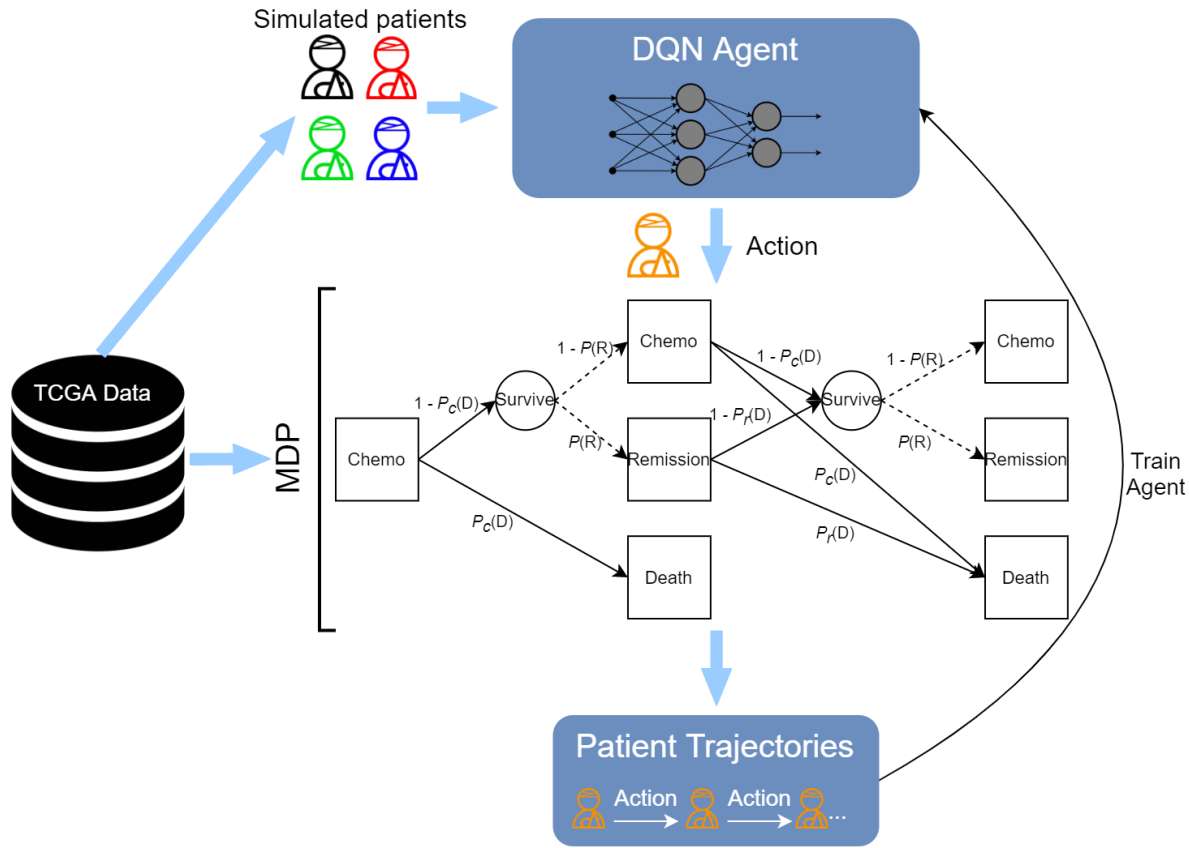

**B**

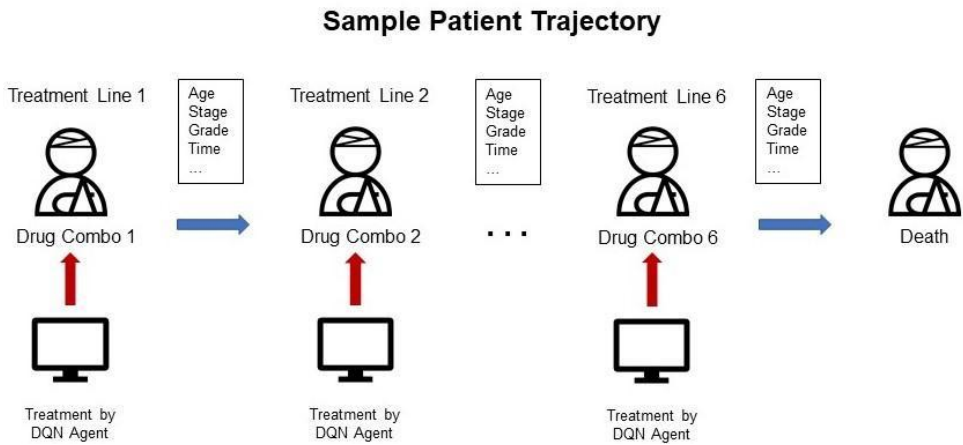

**Figure 1**:(**A**) Overall training schema describing reinforcement learning simulation. TCGA data is used to calculate MDP transition probabilities and sample simulated patients. The DQN agent takes simulated patients (current state) as inputs and produces a drug combination (action) for each patient. The MDP stochastically determines each subsequent state based on state-action pairs from the agent. Patient trajectories are stored in replay memory when the patient reaches a terminal state. Trajectories from previous rounds are used to train the agent. (**B**) Conceptual figure showing the underlying premise of a patient being on a trajectory of care and being treated at each point by a machine.

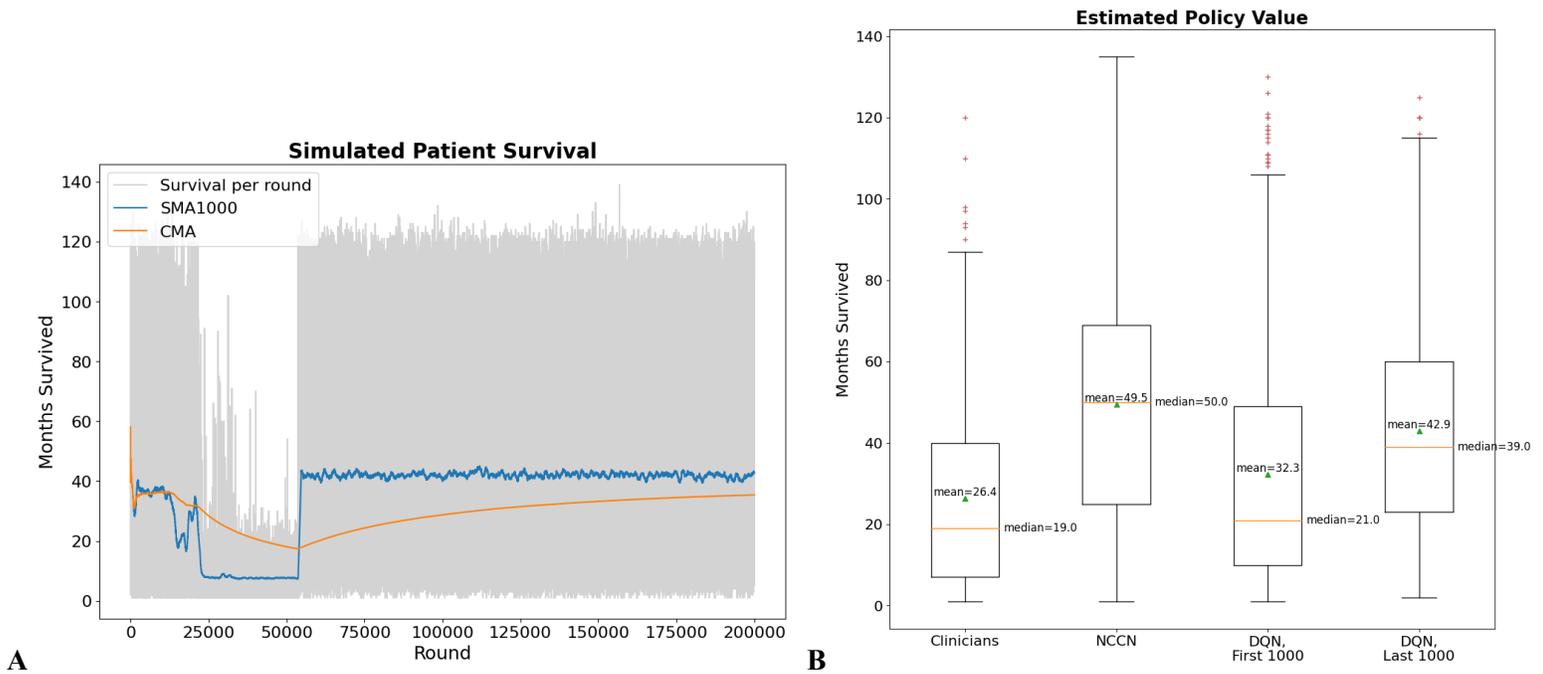

**Figure 2**: (**A**) Training run demonstrating agent learning over 200,000 simulated cases. There is a marked shift in policy after approximately 50,000 patients, whereby the agent learned to use aldesleukin resulting in an increase in mean survival. SMA1000 (blue line) is a simple moving average survival for the previous 1,000 patients, and CMA (orange line) is the cumulative average for all previous patients. (**B**) A box plot demonstrating a comparison between the AI strategies after 200,000 rounds of training. The DQN was able to significantly improve average survival, with a mean survival of 42.9 months for the last 1,000 simulated patients compared to the first 1,000 baseline of 32.3 months ($p<0.000$). The DQN policy also exceeded average survival of patients treated by clinicians (26.4 months, $p<0.000$). The NCCN policy achieved the highest average survival at 49.5 months ($p<0.000$ for clinicians, DQN first 1,000 and DQN last 1,000). Orange lines indicate the median of the series; green triangles indicate the mean; bottom and top box edges indicate the first and third quartiles, respectively; whiskers indicate 1.5 times the interquartile range; red plusses indicate points outside the whisker range.

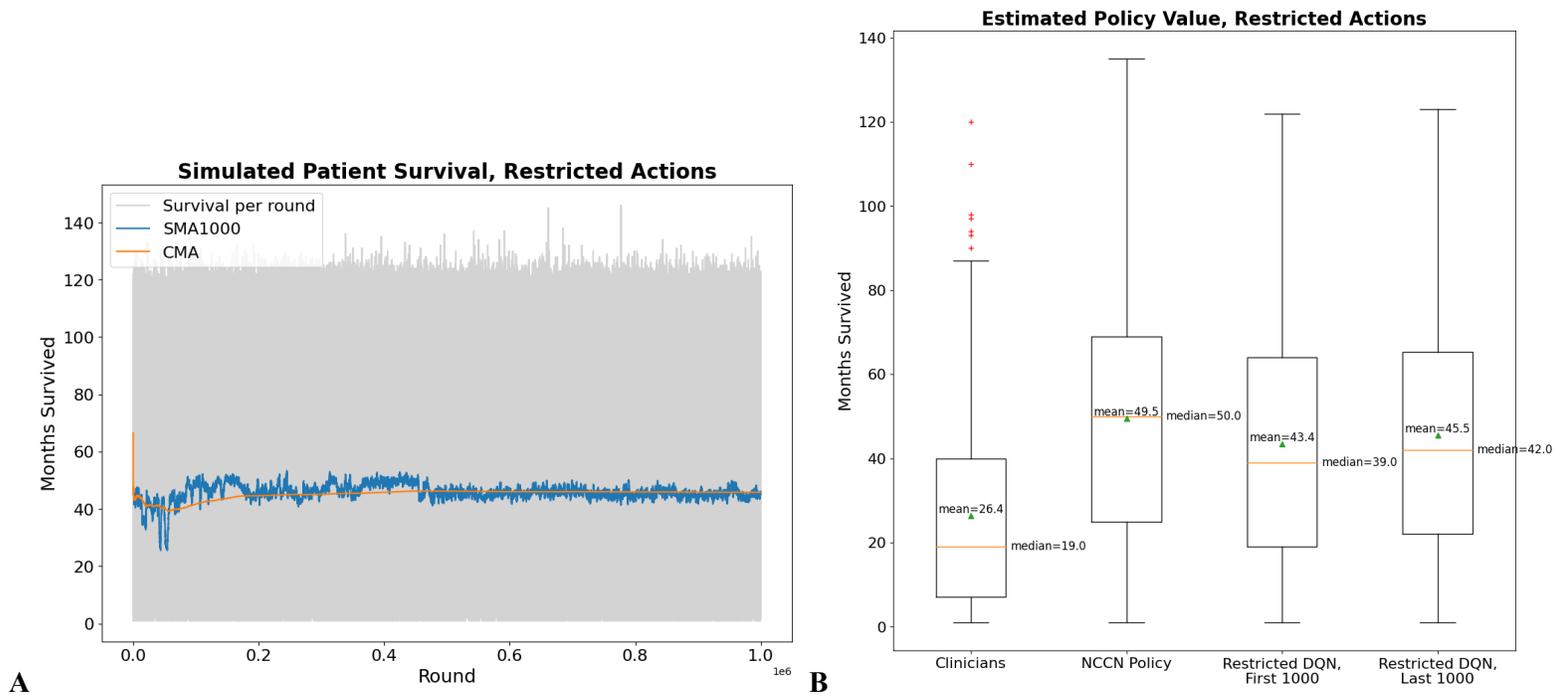

**Figure 3**: We restricted the set of actions of the DQN to exclude uncommon drugs (see Methods) and noted that (**A**) there was a slow but steady increase in the performance of the DQN over the course of training. (**B**) After 1,000,000 rounds of training, the restricted DQN was able to improve average survival, with a mean survival of 45.5 months for the last 1,000 simulated patients compared to the first 1,000 restricted baseline of 43.4 months (p=0.099), and 26.4 months for clinicians (p<0.000). The NCCN policy achieved the highest average survival at 49.5 months (p<0.000 for clinicians, restricted DQN first 1,000 and restricted DQN last 1,000)

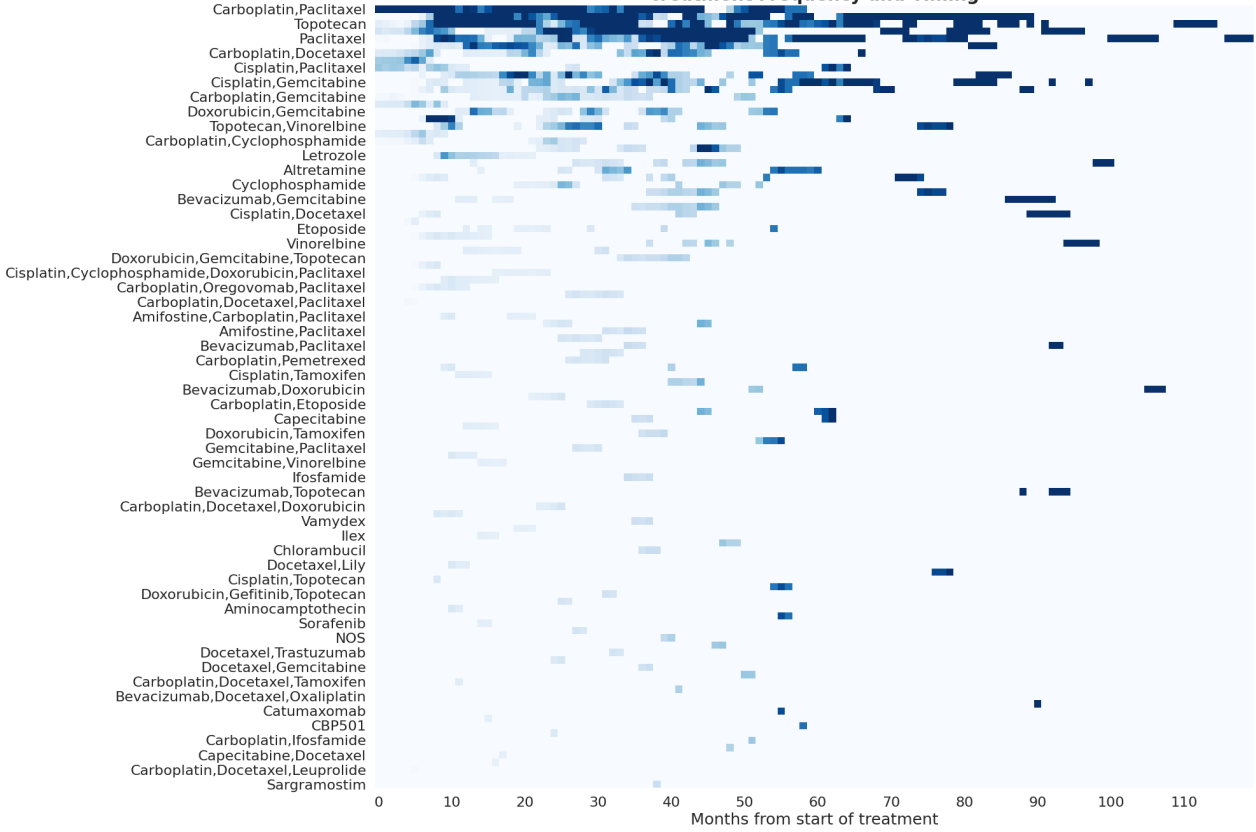

A

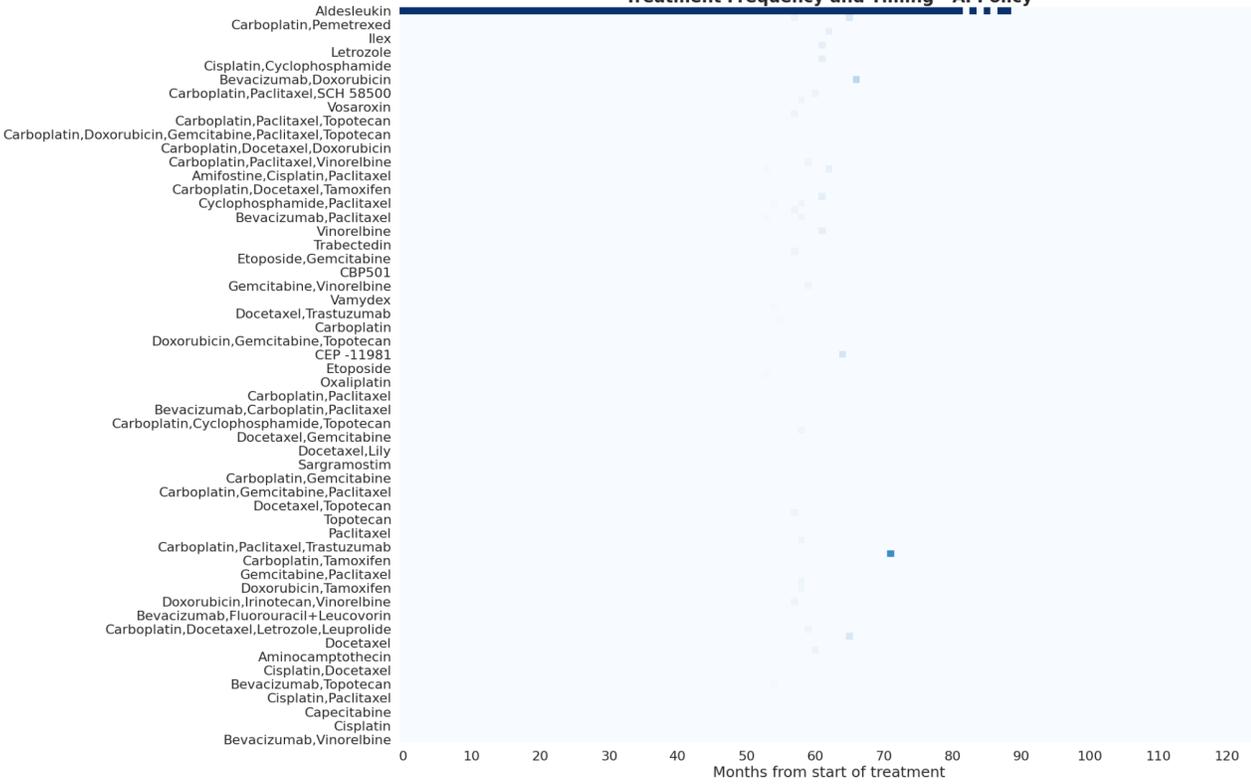

B

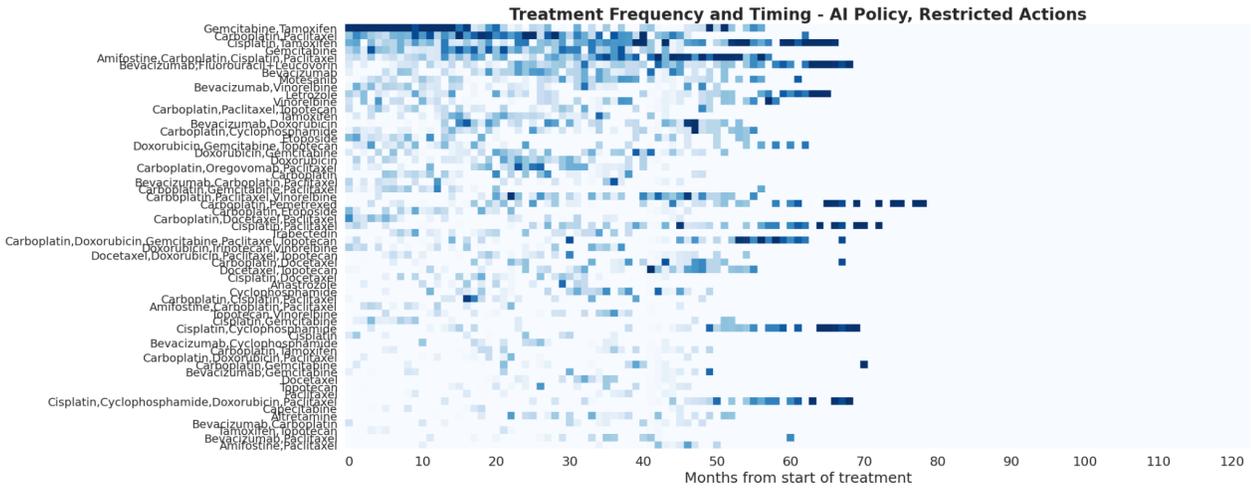

**C**

**Figure 4**: Heatmaps of chemotherapy administration frequencies were normalized from time of initial treatment and converted to z-scores across all drug combinations. Z-scores were subsequently bounded between 0-3 to generate smooth visualizations of drug selection and timing in courses of care by the (**A**) human oncologists, (**B**) DQN, and (**C**) Restricted DQN. The y-axes are sorted by overall frequency per treatment across all time intervals.

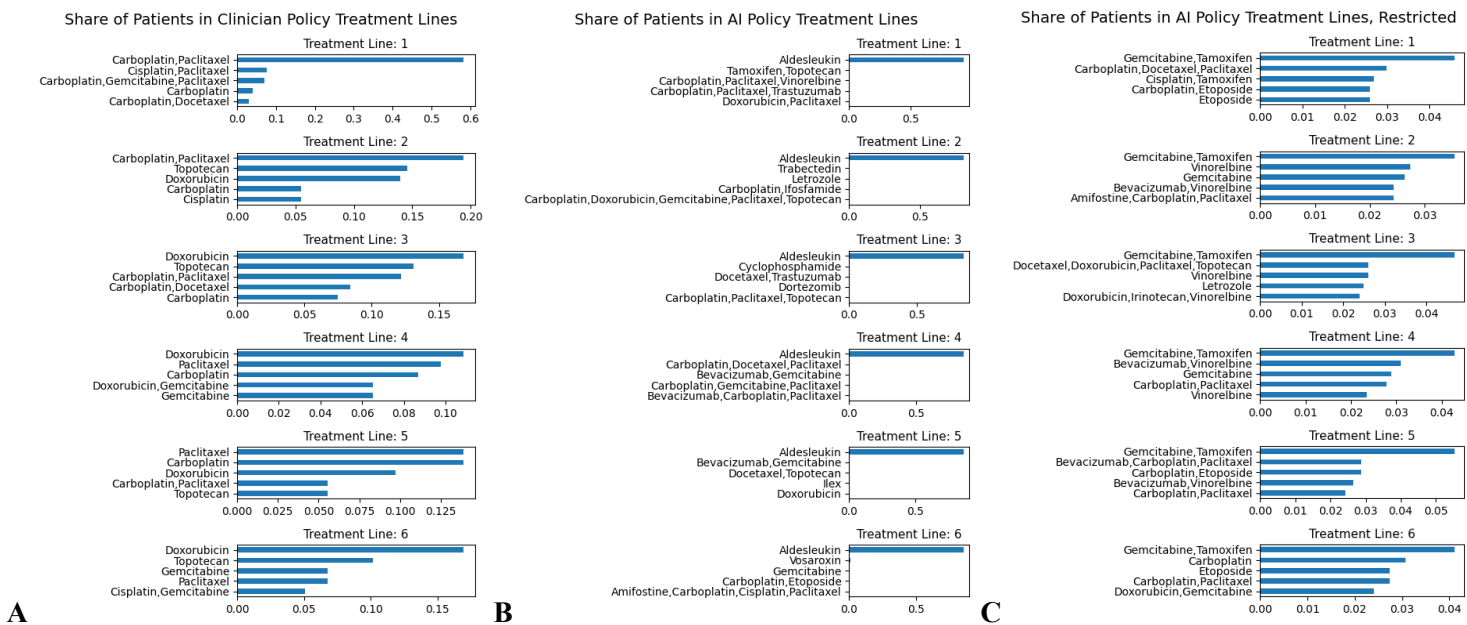

**Figure 5**: Frequency of therapies chosen for each line of treatment by (**A**) Clinicians, (**B**) DQN, (**C**) Restricted DQN. Clinician treatment regimens mirror NCCN guidelines for ovarian cancer. Without restrictions, the DQN favors the use of Aldesleukin almost exclusively, while when limited to common agents, the Restricted DQN tends to favor up-front gemcitabine/tamoxifen, followed by carboplatin/docetaxel/paclitaxel, but doesn't learn a single dominant strategy in later rounds.

**SUPPLEMENTAL FIGURES**

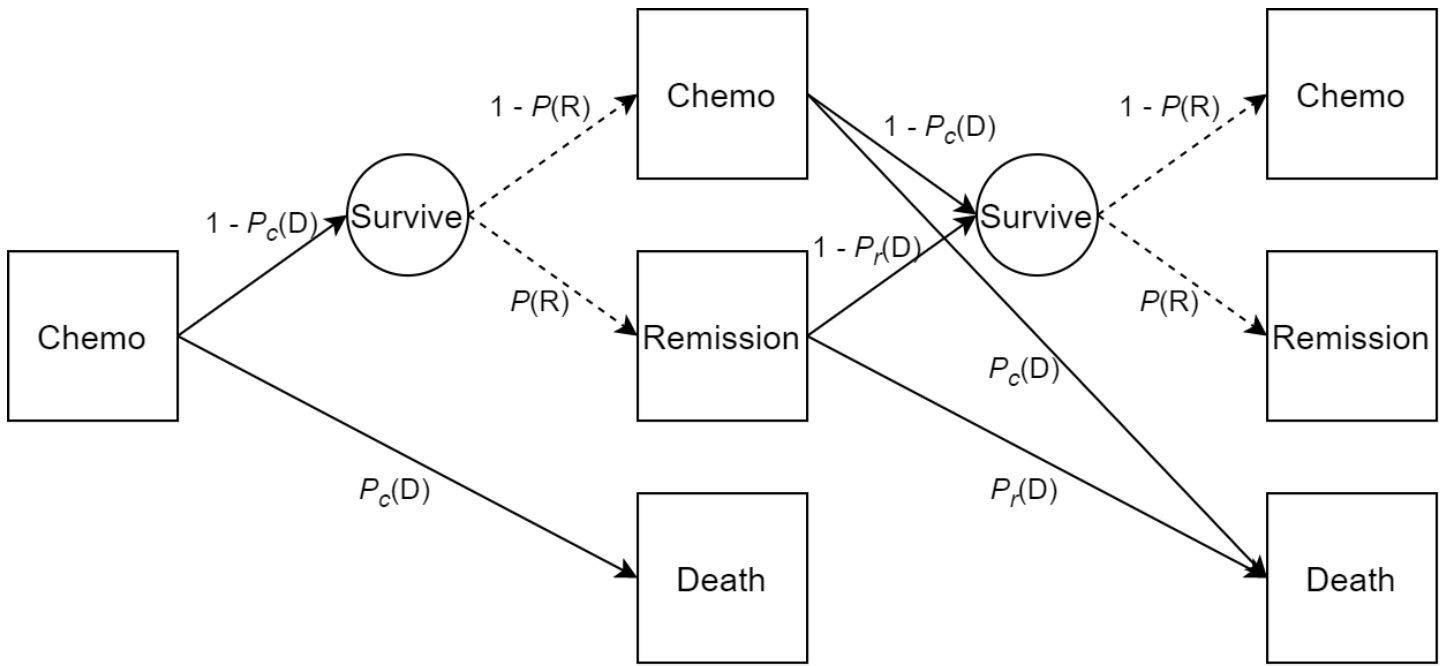

**Supplemental Figure S1**: Markov decision process states and transition probabilities. All patients are assumed to be alive and in need of treatment ("Chemo") in the initial state. Two Cox regressions stochastically determine each subsequent state. The first regression produces a probability of death $P(D)$ from its survival function, and the second, if the patient survives, produces the probability $P(R)$ that the patient will no longer need treatment ("Remission"). Solid arrows denote transition probabilities from the death event regression. Dashed arrows denote probabilities from the recurrence/remission regression. Squares represent states that are stored in memory during the simulation and shown to the agent. "Survive" circles are intermediate states that are not separately recorded, but determine whether the recurrence/remission regression probabilities are used.

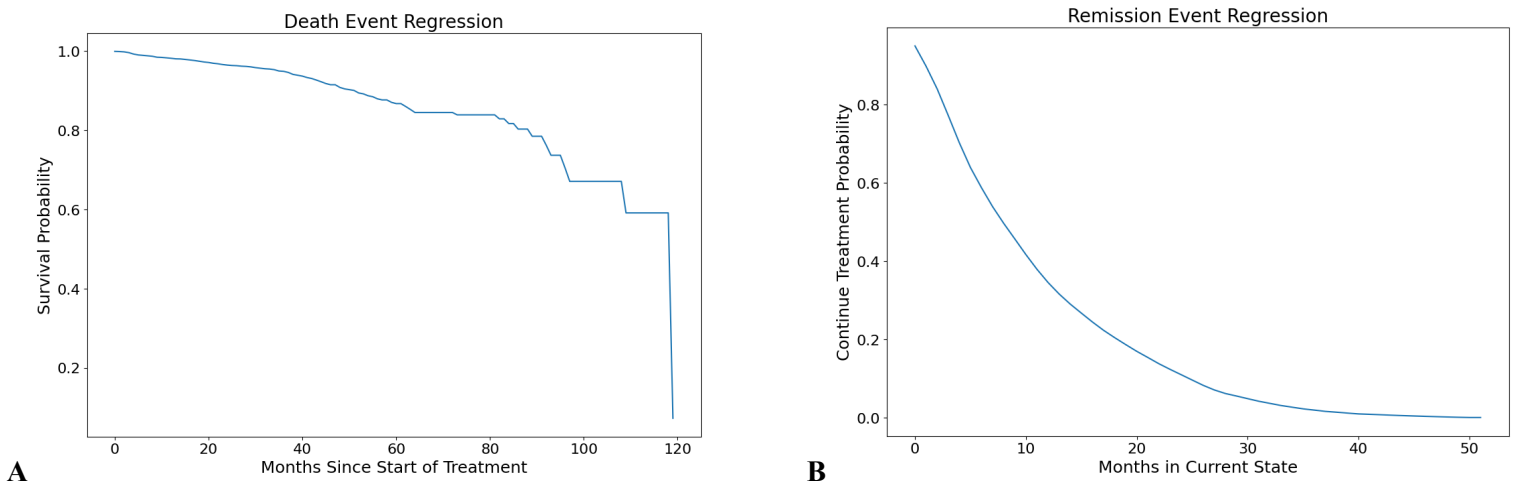

**Supplemental Figure S2**: Baseline survival curves from Cox-proportional hazard modeling of (**A**) death and (**B**) recurrence.

## SUPPLEMENTAL METHODS

### Reinforcement Learning Model
*Environment*

Using the reorganized TCGA data, we constructed a Markov decision process (MDP) to simulate the treatment decisions and survival trajectories of epithelial ovarian cancer patients.(7,14,17) This served as the reinforcement learning environment in the model. The environment consists of four main components: $S$, $A$, $T(s', s, a)$, and $R(s')$.

$S$ is a finite set of states that describe the health status of the patient. Each state $s$ in $S$ includes whether the patient is alive and needs treatment, is alive with cancer in remission, or is deceased, (health state); months since the start of treatment, defined as the number of previous states the patient has survived; the number of consecutive previous states that include the current treatment; age at the start of treatment; race; and tumor stage and grade at the start of treatment. Age, race, tumor stage, and tumor grade are assigned to simulated patients by sampling values based on their frequency in the TCGA data.

$A$ is a finite set of actions that can be chosen by a reinforcement learning agent. $A$ consists of all the unique treatment combinations in the reorganized TCGA drug lines data, and each action $a$ in $A$ is a drug or drug combination. $A$ also includes a no-treatment option, but the agent must choose this action if the current state includes a no-treatment-needed health state, and the agent cannot choose it if the current state includes a treatment-needed health state. This mechanic is designed to avoid conflating the *state* of not needing cancer treatment with the *action* of taking the patient off all treatments, which by definition would lead to a no-treatment health state. Thus, the agent is not tasked with diagnosing the status of the patient's cancer, but only with selecting a drug regimen when the environment determines that the patient requires treatment. Theoretically, a clinician or agent could select any therapeutic combination that could be constructed from all reasonable choices of chemotherapeutic agents. However, we limited $A$ to include only those combinations present in the TCGA ovarian cancer dataset. We did this to prevent the agent from selecting drug combinations that are not in the data because our model cannot calculate transition probabilities for these treatments.

$T(s', s, a)$ is a set of transition probabilities giving the likelihood that an action $a$ chosen in state $s$ at time $t$ will result in the next state $s'$ at time $t + 1$.(14,17) Each state transition involves two sets of probabilities that stochastically determine the subsequent state. The first determines whether patients die with probability $P(D)$ or survive with probability $P(S) = 1 - P(D)$. If patients die, the subsequent state is death, the terminal state, and simulation proceeds to the next round (i.e., a new simulated patient). If patients survive, the next set of probabilities determines whether their cancer enters remission, $P(R)$, or requires further treatment in the next state, $P(T) = 1 - P(R)$. Supplemental Figure S1 illustrates the interaction of the transition probabilities in $T(s', s, a)$ with the health state for each state $s$.

To calculate these probabilities, we used a survival analysis based on the Cox proportional hazard regression model.(33) Survival analyses often use Cox models to investigate the time to the occurrence of events and the probability of an event's occurrence at a given time interval in the form of a hazard statistic, $\lambda$.(34) Using the *lifelines* python package (https://lifelines.readthedocs.io/en/latest/index.html), we implemented two separate Cox regressions to calculate the transition probabilities for $T(s', s, a)$. The first, the death event regression, is a common Cox model, and estimates the likelihood of a patient dying (a terminal event) at one-month time intervals from the start of treatment.(33,34) The following equation gives the death event regression hazard for a patient $i$:

$$\lambda_i(t) = \lambda_0(t) exp(\beta X_i), \ i = 1,..., n,$$

where $\lambda_0(t)$ is the baseline hazard, accounting for the likelihood of an event with respect to time $t$ across all $i$, and $exp(\beta X_i)$ is the partial or treatment hazard, which is constant over $t$ and accounts for all other features specific to each $i$.(33,34) $\beta$ is a vector of coefficients corresponding to $X_i$, which includes features for each patient's current health state, current treatment, number of previous lines of treatment received, age, race, tumor stage, and tumor grade.(29,33,34) $P(D)$ and $P(S)$ are determined by sampling from the survival function of $\lambda_i(t)$ according to the patient's current state-action pair. The survival function gives the probability that a patient has not experienced an event at time $t$, and is defined as (35):

$$S(t) = exp(-\int_0^t \lambda(x)dx)$$

The second model, the recurrence/remission regression, is based on the Prentice-Williams-Peterson approach to the Cox model, which assumes that the event in question is recurrent and considers the number of past occurrences of the

event.(34,36) We use the gap-time version of the Prentice-Williams-Peterson model, which uses time since the previous event to calculate the baseline hazard.(34,36,37) The following equation gives the recurrence/remission regression hazard for a patient $i$ and recurrent event $j$(34,36):

$$\lambda_{ij}(t) = \lambda_{0j}(t - t_{j-1})exp(\beta_j X_{ij}), \ i = 1,..., n; \ j = 1,..., k_i, \ k \leq k$$

$X_{ij}$ includes features for each patient's current health state, current treatment, months since start of treatment, number of previous lines of treatment received, age, race, tumor stage, and tumor grade. $P(R)$ and $P(T)$ are determined by sampling from the survival function of $\lambda_{ij}(t)$ according to the patient's current state-action pair. We used the *lifelines* CoxPHFitter class to implement both regressions with a penalizer term of 0.1 to control for high correlation between covariates (see, https://lifelines.readthedocs.io/en/latest/fitters/regression/CoxPHFitter.html#lifelines.fitters.coxph_fitter.SemiParametricPHFitter.predict_cumulative_hazard). Supplemental Figures S2A and S2B show the baseline survival curves for the death event and recurrence/remission regressions, respectively.

$R(s')$ is the immediate reward administered to the agent by the environment when it transitions to the subsequent state $s'$. In this study, $R(s')$ is simply +1 if $s'$ is not death and -1 if $s'$ is death, and is administered to the agent after each state transition. Thus, the total reward after one round of the simulation is the total number of months of patient survival where the action $a$ did not result in death in the next state.

The environment starts by initializing a "patient" in the needs-treatment health state at time $t = 0$ with all other state features randomly generated based on the TCGA data and held fixed throughout the round of the simulation. The agent then selects an action, which along with the state, determines the probabilities of the treatment status of the patient in the subsequent state (needs treatment, does not need treatment, deceased). The next state is then determined stochastically from these probabilities, and a reward signal is administered and added to the cumulative reward. Additionally, the state transition is stored for use by a policy in selecting actions in subsequent rounds. This process is repeated until the environment transitions to the "death" state, and the total reward for the round and the trajectory of treatments and transitions (patient trajectory) is stored. The environment runs for a specified number of rounds, each with one patient being treated by the agent, and produces the total reward and full treatment trajectory for each round.

*Policy*

When in a needs-treatment health state, the agent selects actions by applying a policy π to the stored state transitions, which compose the replay memory. During each state, the policy trains a model on this replay memory to predict the optimal action to maximize total reward. The policy includes an exploration rate $\epsilon \in [0, 1]$, which is the probability that the agent will choose an action randomly. Thus, the policy is an ϵ-greedy policy.(14) ϵ starts at 0.9 and decays after each round until it reaches 0.05. The policy is a deep Q-network (DQN) in the form of a multi-layer perceptron neural network with six fully connected hidden layers, ReLU activation functions, and a fully connected output layer that generates a probability that each action is the optimal action. The DQN policy is optimized after each patient trajectory using off-policy temporal difference (TD) learning. This optimization calculates loss based on the expected state-action values $Q^\pi(s, a)$ and the actual state-action values from saved trajectories. $Q^\pi(s, a)$ is updated using the following equation (14,17):

$$Q^\pi(s, a) \leftarrow Q^\pi(s, a) + \alpha \cdot [r + \gamma \cdot Q^\pi(s', a') - Q^\pi(s, a)], \ for \ all \ s \in S, \ a \in A(s)$$

$Q^\pi(s', a')$ is determined by a target policy model that uses the DQN weights from a previous round to calculate the action values. $\gamma \in [0, 1]$ is the discount rate, which determines how much future rewards are valued relative to immediate rewards. This tells the agent how much it should value future survival relative to survival in the immediate next state.(14) We used a discount rate $\gamma = 0.99$ in this analysis. $\alpha$ is a constant step-size parameter (14) set at 0.01 in this analysis. We use the RMSprop algorithm and smooth L1 loss to optimize the DQN policy, implemented with pytorch.(38,39)

In addition to the DQN agent, we developed an agent based on the National Comprehensive Cancer Network (NCCN) guidelines for ovarian cancer treatment to choose the drug combinations for the simulated patient at each time step.(9) Specifically, the NCCN agent is a rules-based policy where we mapped the "preferred regimens" and "other recommended regimens" noted in the NCCN guidelines' Principles of Systemic Therapy of Ovarian Cancer to the drug combinations present in our dataset.(9) In treating a simulated patient, the NCCN agent randomly selects one of the "preferred regimens" for the given disease stage. If after using the "preferred regimens," the cancer for the simulated patient recurs, then the NCCN agent randomly selects one of the "other recommended regimens" for the given disease stage. We only allowed the NCCN agent to choose actions that are listed in the NCCN guidelines and that occur at least once in the TCGA data.

Figure 1 summarizes the structure of the entire reinforcement learning simulation. This begins with the reorganized TCGA dataset, which we use to create the MDP components: the state (simulated patients and their health state), actions (unique drug combinations), and transition probabilities. The environment shows simulated patients to the DQN agent, which selects an action in each state. Each state-action pair determines the transition probabilities for survival and subsequent health state until the simulated patient reaches the terminal state. The entire patient trajectory is saved into replay memory, and the DQN optimizes based on the replay memory. This process repeats for a specified number of rounds.